\def\year{2020}\relax
\documentclass[letterpaper]{article} 
\usepackage{aaai20}  
\usepackage{times}  
\usepackage{helvet} 
\usepackage{courier}  
\usepackage[hyphens]{url}  
\usepackage{graphicx} 
\usepackage{graphicx}  
\usepackage{mathrsfs}
\usepackage{amsfonts}
\usepackage{amsmath}
\usepackage{multirow}
\usepackage{bm}
\usepackage{booktabs}
\usepackage{array}
\usepackage[linesnumbered,ruled,lined]{algorithm2e}
\usepackage{algpseudocode}
\title{Semi-Supervised Text Simplification with Back-Translation and Asymmetric Denoising Autoencoders}
\author{Yanbin Zhao, Lu Chen, Zhi Chen, Kai Yu\thanks{Kai Yu is the corresponding author.}\\
MoE Key Lab of Artificial Intelligence\\
SpeechLab, Department of Computer Science and Engineering\\
Shanghai Jiao Tong University, Shanghai, China\\
\{zhaoyb, chenlusz, zhenchi713, kai.yu\}@sjtu.edu.cn
}
 
\begin{document}

\maketitle

\begin{abstract}
Text simplification (TS) rephrases long sentences into simplified variants while preserving inherent semantics. Traditional sequence-to-sequence models heavily rely on the quantity and quality of parallel sentences, which limits their applicability in different languages and domains. This work investigates how to leverage large amounts of unpaired corpora in TS task. We adopt the back-translation architecture in unsupervised machine translation (NMT), including denoising autoencoders for language modeling and automatic generation of parallel data by iterative back-translation. However, it is non-trivial to generate appropriate complex-simple pair if we directly treat the set of simple and complex corpora as two different languages, since the two types of sentences are quite similar and it is hard for the model to capture the characteristics in different types of sentences. To tackle this problem, we propose asymmetric denoising methods for sentences with separate complexity. When modeling simple and complex sentences with autoencoders, we introduce different types of noise into the training process. Such a method can significantly improve the simplification performance. Our model can be trained in both unsupervised and semi-supervised manner. Automatic and human evaluations show that our unsupervised model outperforms the previous systems, and with limited supervision,  our model can perform competitively with multiple state-of-the-art simplification systems.
\end{abstract}

\section{Introduction}
\label{sec:intro}
Text simplification reduces the complexity of a sentence in both lexical and structural aspects in order to increase its intelligibility. It brings benefits to individuals with low language skills and has abundant usage scenarios in education and journalism fields~\cite{de2010text}. Also, a simplified version of a text is easier to process for downstream tasks, such as parsing~\cite{chandrasekar1996motivations}, semantic role labeling~\cite{woodsend2011learning}, and information
extraction~\cite{jonnalagadda2010biosimplify}.

Most of the prior works regard this task as a monolingual machine translation problem and utilize sequence-to-sequence architecture to model this process~\cite{nisioi-etal-2017-exploring,zhang-lapata-2017-sentence}. These systems rely on large corpus containing pairs of complex and simplified sentences, which severely restrict their usage in different languages and the adaptation to downstream tasks in different domains. So, it is essential to explore unsupervised or semi-supervised learning paradigm which can effectively work with unpaired data.

In this work, we adopt back-translation~\cite{DBLP:conf/acl/SennrichHB16} framework to perform unsupervised and semi-supervised text simplification. Back-translation converts the unsupervised task into a supervised one by on-the-fly sentence pair generation. It has been successfully used in unsupervised neural machine translation~\cite{DBLP:conf/iclr/ArtetxeLAC18,DBLP:conf/emnlp/LampleOCDR18}, semantic parsing~\cite{cao-etal-2019-semantic} and natural language understanding~\cite{zjz17-zhao-emnlp19}. Denoising autoencoder (DAE)~\cite{DBLP:conf/icml/VincentLBM08} plays an essential part in back-translation model. It performs language modeling and helps the system learn useful structures and features from the monolingual data. In NMT task, the translations between different languages are equal, and the denoising autoencoders have a symmetric structure, which means different languages use the same types of noise (mainly word dropout and shuffle). However, if we treat the set of simple and complex sentences as two different languages, the translation processes are asymmetric: Translation from simple to complex is an extension process requires extra generations, while information distillation is needed during the inverse translation. Moreover, text simplification is a monolingual translation task. The inputs and outputs are quite similar, which makes it more difficult to capture the different features in complex and simple sentences. As a result, symmetric denoising autoencoders may not very helpful in modeling sentences with diverse complexity and make it non-trivial to generate appropriate parallel data.

To tackle this problem, we propose asymmetric denoising autoencoders for sentences with different complexity. We analyze the effects of denoising type on the simplification performance and show that separate denoising methods is beneficial for decoders to generate suitable sentences with different complexity. Besides, we set several criteria to evaluate the generated sentences and use policy gradient to optimize these metrics. We use this as an additional method to improve the quality of the generated sentences. Our approach relies on two unpaired corpora -- one is statistically simpler than another. In summary, our contributions include:
\begin{itemize}
  \item We adopt the back-translation framework to utilize large amounts of unpaired sentences for text simplification.
  \item We propose asymmetric denoising autoencoders for sentences with different complexity and analyze the corresponding effects.
  \item We develop methods to evaluate both simple and complex sentences derived from back-translation and use reinforcement learning algorithms to promote the quality of the back-translated sentences.
\end{itemize}

\section{Related Works}
As a monolingual translation task, early text simplification systems usually based on statistical machine translation such as PBMT-R~\cite{wubben2012sentence} and Hybrid~\cite{narayan2014hybrid}. \citeauthor{xu2016optimizing}~\shortcite{xu2016optimizing} achieved state-of-the-art performance by leveraging paraphrases rules
extracted from bilingual texts. Recently, neural network models have been widely used in simplification systems. \citeauthor{nisioi-etal-2017-exploring}~\shortcite{nisioi-etal-2017-exploring} first applied Seq2Seq architecture to model text simplification. Several extensions are also proposed for this architecture such as augmented memory~\cite{vu-etal-2018-sentence} and multi-task learning~\cite{DBLP:conf/coling/GuoPB18}. Furthermore,
 \citeauthor{zhang-lapata-2017-sentence}~\shortcite{zhang-lapata-2017-sentence} proposed DRESS, a Seq2Seq model trained in a reinforcement learning framework. Sentences with high fluency, simplicity and adequacy are rewarded during the training process. ~\citeauthor{zhao-etal-2018-integrating}~\shortcite{zhao-etal-2018-integrating} utilized Transformer~\cite{vaswani2017attention} integrated with external knowledge and achieved state-of-the-art performance in automatic evaluation. \citeauthor{Complexity-Weighted}~\shortcite{Complexity-Weighted} proposed complexity-weighted loss and a reranking system to improve the simplicity of the sentences. Systems all above require large amounts of paralleled data.

 In terms of unsupervised simplification, several systems only perform lexical simplification~\cite{DBLP:conf/inlg/NarayanG16,paetzold2016unsupervised} by replacing complicated words with their simpler synonyms, which ignored other operations such as reordering and rephrasing. \citeauthor{unsuper-nerual-simplificaiton}~\shortcite{unsuper-nerual-simplificaiton} proposed an unsupervised method for neural models. They utilized adversarial training to enforce a similar attention distribution between complex and simple sentences. They also tried back-translation with normal denoising techniques but did not achieve preferable results. We think it is inappropriate to apply back-translation framework mechanically into simplification task. So in this work, we make several improvements and achieve promising results.
\section{Our Approach}
\subsection{Overview}
\begin{figure*}[tb]
    \centering
    \includegraphics[width=0.85\textwidth]{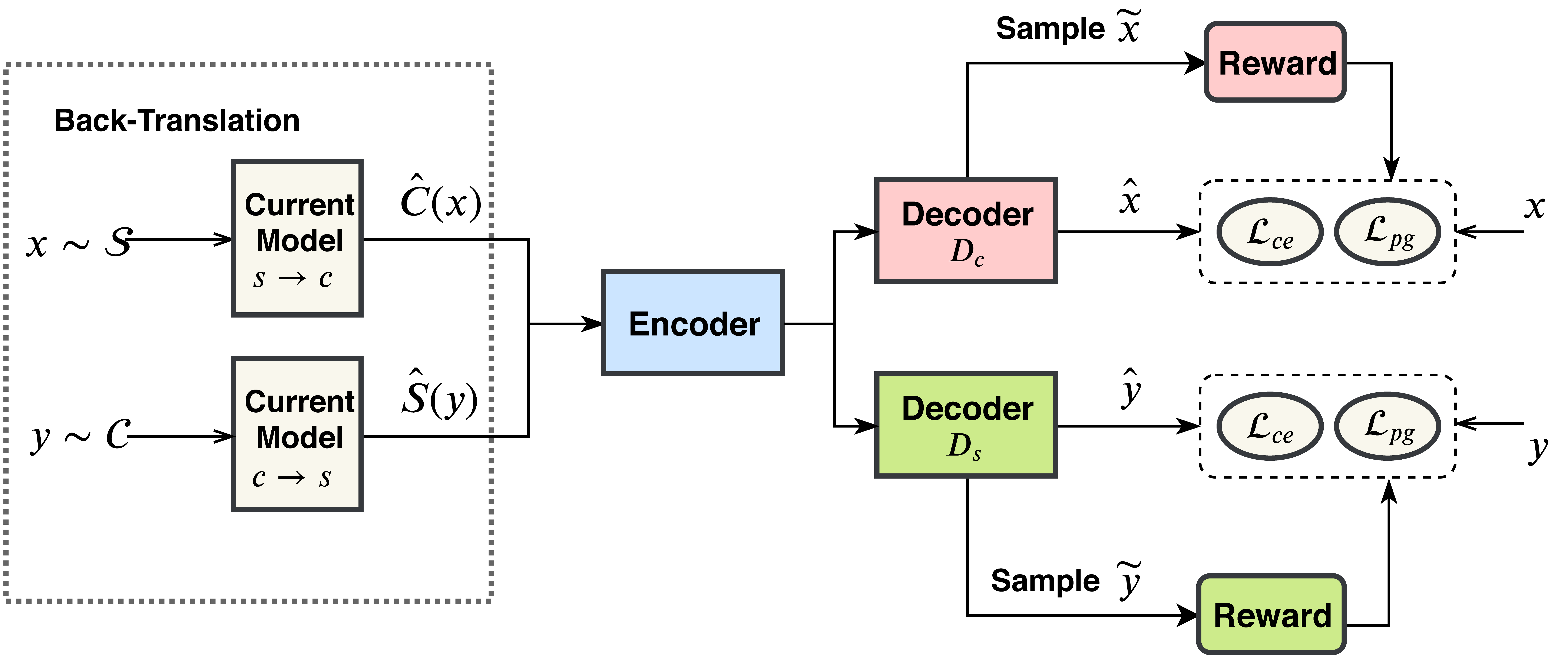}
    \caption{Overview of the our proposed system. Back-translated sentences $\hat{C}(x)$, $\hat{S}(y)$ and their original inputs $x$, $y$ form sentences pairs. ($\hat{C}(x)$, $x$) is used to train complex-simple model, and ($\hat{S}(y)$, $y$) is used to train simple-complex model. The model parameters are updated with both cross entropy loss and policy gradient.}
    \label{fig:overview}
\end{figure*}

The architecture of our simplification system is illustrated in Figure \ref{fig:overview}. The system consists of a shared encoder $E$ and a pair of independent decoders: $D_s$ for simple sentences and $D_c$ for complex sentences. Denote the corresponding sentence spaces by $\mathcal{S}$ and $\mathcal{C}$. The encoder and decoders are first pre-trained as asymmetric denoising autoencoders (See below) on separated data. Next, the model goes through an iterative process. At each iteration, simple sentence $x \in \mathcal{S}$ is translated to a relatively complicated one $\hat{C}(x)$ via current model $E$ and $D_c$. Similarly, complex sentence $y \in \mathcal{C}$ is translated to a relatively simple version $\hat{S}(y)$ via $E$ and $D_s$. The pairs $(\hat{C}(x), x)$ and $(\hat{S}(y), y)$ are automatically-generated parallel sentences which can be used to train the model in a supervised manner with cross entropy loss. During the supervised training, our current model can also be regarded as translation policies. Let $\widetilde{x}$, $\widetilde{y}$ denote the simple and complex sentences sampled from the current policies. Corresponding rewards $R_s$ and $R_c$ is calculated according to their quality. The model parameters are updated with both cross entropy loss and policy gradient.

\subsection{Back-Translation Framework}
In the back-translation framework, the shared encoder aims to represent both simple and complex sentences in a same latent space, and the decoders need to decompose this representation into sentences with corresponding types.
We update the model by minimizing the cross entropy loss:
\begin{equation}
    \begin{split}
        \mathcal{L}_{ce} = & \mathbb{E}_{x \sim \mathcal{S}} \left [-\log P_{c \to s}(x |\hat{C}(x))\right] + \\
        & \mathbb{E}_{y \sim \mathcal{C}} \left [-\log P_{s \to c}(y |\hat{S}(y))\right]
    \end{split}
    \label{eq:back-translation}
\end{equation}
Where $P_{c \to s}$ and $P_{s \to c}$ represent the translation models from complex to simple and vice versa. The updated model tends to generate better synthetic sentence pairs for the next training process. Through such iterations, the model and back-translation process can promote mutually and finally lead to a good performance.
\subsection{Denoising}
\label{ssec:noise}
\citeauthor{LampleCDR18}~\shortcite{LampleCDR18} showed that denoising strategies such as word dropout and shuffle have a critical impact on unsupervised NMT systems. We argue that these symmetric noises in NMT may not be very effective in simplification task. So in this section, we will describe our asymmetric noises for simple and complex corpus.

\subsubsection{Noise for Simple Sentences}
Sentence with low complexity tends to have simple words and structures. We introduce three types of noise to help the model capture these characteristics.
\\\\
\textbf{Substitution}: We replace the relatively simple words into advanced expressions with the guidance of Simple PPDB~\cite{DBLP:conf/acl/PavlickC16a}. Simple PPDB is a subset of the Paraphrase Database (PPDB)~\cite{DBLP:conf/naacl/GanitkevitchDC13} adapted for the simplification task. It contains 4.5 million pairs of complex and simplified phrase. Each pair constitutes a simplification rule and has a score to indicate the confidence.

Table \ref{tab:simpleppdb} shows several examples, where advance expression such as ``fatigued'' and ``wary'' can be simplified to ``tired''. However, in this situation, we utilize these rules in the reverse direction, meaning if ``tired'' appears in the sentence, it can be replaced by one of the candidates above with probability $P_{rep}$. In our experiments, $P_{rep}$ is set to 0.9. Rules with scores lower than 0.5 will be discarded, and we only choose the top five phrases with the highest confidence score as the candidates for each word. During the substitution process, a substitute expression is randomly sampled from the candidates and replace the original phrases.
\begin{table}[htbp]
    \centering
    \begin{tabular}{|c|l|}
        \hline
        \textbf{Score} & \textbf{Rules}\\
        \hline
        0.95516 & completely exhaust $\to$ tired \\
        \hline
        0.82977 & fatigued $\to$ tired \\
        \hline
        0.79654 & weary $\to$ tired \\
        \hline
        0.57126 & tiring $\to$ tired \\
        \hline
    \end{tabular}
    \caption{Examples from the Simple PPDB}
    \label{tab:simpleppdb}
\end{table}

Substitution helps the model learn words distribution from the simplified sentences. To some extent, it also simulates the lexical simplification process, which can encourage decoder $D_s$ to generate simpler words from the shared latent space.\\\\
\textbf{Additive}: Additive noise inserts additional words into the input sentences.  \citeauthor{fevry2018unsupervised}~\shortcite{fevry2018unsupervised} used autoencoder with additive noise to perform sentence compression, and generate imperfect but valid sentence summaries. Additive noise forces the model to subsample words from the corrupt inputs and generate reasonable sentences. It can help the model capture sentence trunk in simplification task.

For an original input, we randomly select another sentence from the training set and sample a subsequence without replacement. We then insert the subsequence to the original input. Instead of sampling independent words, we sample bigrams from the additional sentence. The subsequence length depends on the length of the original input. In our experiments, the sampled sequence serving as noise accounts for 25\%-35\% of the whole noised sentence.\\\\
\textbf{Shuffle}: Word shuffling is a common noising method in denoising autoencoders. It is proven to be helpful for the model to learn useful structure in sentences~\cite{LampleCDR18}. To make the additive words evenly distributed in the noised simple sentence, we concatenate the original sentence and the additive subsequence and \emph{complete} shuffle the bigrams, keeping all word pairs together.

An example noising process on simple sentence is illustrated in Table \ref{tab:denoising_example}.
\begin{table}[htbp]
    \centering
    \begin{tabular}{|l|m{0.5\columnwidth}|}
         \hline
         Original & Their voices sound tired \\
         \hline
          + substitution & Their voices sound \textbf{exhausted}\\
         \hline
          \quad + additive \& shuffle & sound \textbf{exhausted} \emph{he knows} Their voices  \\
         \hline
    \end{tabular}
    \caption{Example of the noising process on simple sentences. The italic words is the additive noise sampled from another sentence.}
    \label{tab:denoising_example}
\end{table}
\subsubsection{Noise for Complex Sentences}
Substitution is also performed for complex sentences. Here, we use the rules in Simple PPDB normally to rewrite the complicated words into their simpler versions. Rest of the process is the same as the substitution method for simple sentences. Apart from this, we applied other two noising methods.\\\\
\textbf{Drop}: Word dropping discards several words from the sentences. During the reconstruction, the decoder has to recover the removed words through the context. Translation from simple to complex usually include sentence expansion, which needs the decoder to generate extra words and phrases. Word dropping can align autoencoding task closer with sentence expansion and promote the quality of the generated sentences.

Since words with lower frequency usually contain more semantic information,  we only delete the ``frequent word'' with probability $P_{del}$. We define ``frequent word'' as the word with more than 100 occurrences in the entire corpus. A similar approach has also been used in unsupervised language generation~\cite{freitag2018unsupervised}. We set $P_{del} = 0.6$ in our experiments.\\\\
\textbf{Shuffle}: Different from the complete shuffle process for simple sentences, we only \emph{slightly} shuffle the complex sentences. This is because complex sentences don’t have additive noise, and when the sentences get longer and more complex, it is hard for the decoder to reconstruct the sentences with the complete shuffled inputs. Similar to~\citeauthor{LampleCDR18}\shortcite{LampleCDR18}, the max distance $k$ between shuffled word and its original position is limited.
\begin{figure}[htbp]
    \centering
    \includegraphics[width=1\columnwidth]{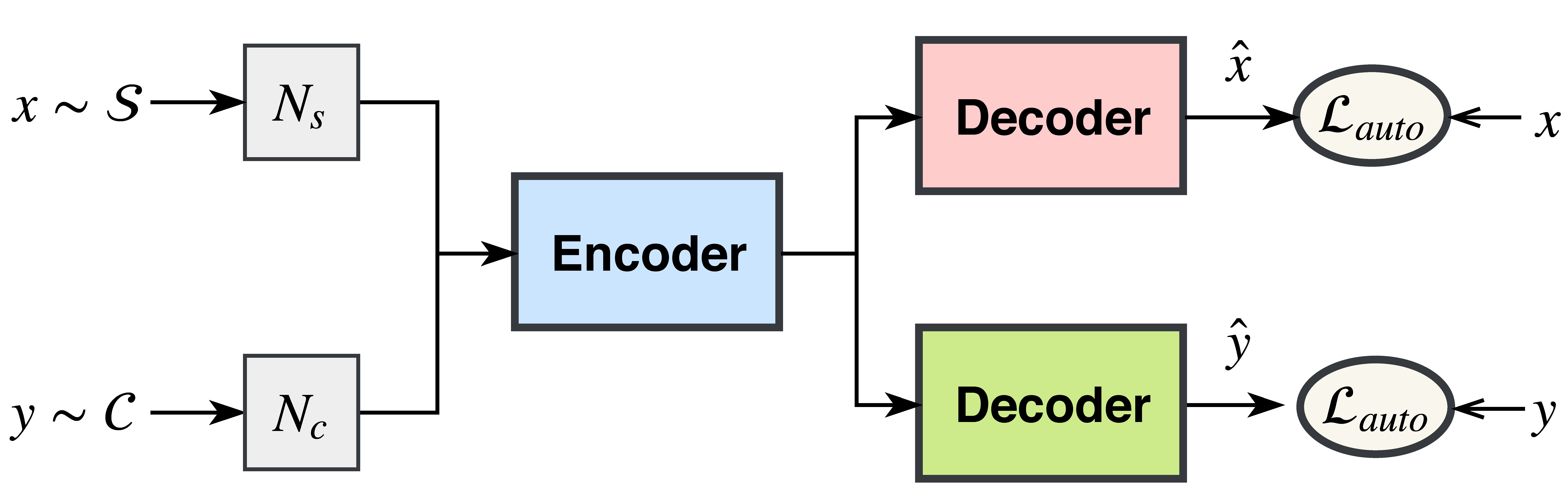}
    \caption{Training process of the asymmetric denoising autoencoders}
    \label{fig:denoising_process}
\end{figure}

We train the denoising autoencoders by minimizing the loss function:
\begin{equation}
    \begin{split}
        \mathcal{L}_{auto} = & \mathbb{E}_{x \sim \mathcal{S}} \left [-\log P_{s \to s}(x |N_s(x))\right] + \\
        & \mathbb{E}_{y \sim \mathcal{C}} \left [-\log P_{c \to c}(y |N_c(y))\right]
    \end{split}
    \label{eq:back-translation}
\end{equation}
Where $N_s$ and $N_c$ is the noising function for simple and complex sentences. $P_{s \to s}$ and $P_{c \to c}$ denote the corresponding autoencoders. Figure \ref{fig:denoising_process} illustrated the training process.
\subsection{Reward in Back-Translation}
In order to further improve the training process and generate more appropriate sentences for the following iterations, we proposed three ranking metrics as the reward and directly optimize these metrics through policy gradient:\\\\
\textbf{Fluency}: The fluency of a sentence is measured by language models. We trained two LSTM language models~\cite{mikolov2010recurrent} for both types of sentences with the corresponding data. For sentence $x$, The fluency reward $r_f$ is derived from its perplexity and scaled to $\left[0 \sim 1\right]$:
\begin{equation*}
    r_f(x) = \exp\left\{\frac{1}{\left |x\right |} \sum_{i=1}^{\left |x \right |}\log P_{LM}(x_i|x_{0:i-1})\right\}
\end{equation*}
\textbf{Relevance}: Relevance score $r_s$ indicate how well the meaning is preserved during the translation. For inputs and sampled sentences, we generate sentence vectors by taking a weighted average of word embeddings~\cite{DBLP:conf/iclr/AroraLM17} and calculate the cosine similarity.\\\\
\textbf{Complexity}: Complexity reward $r_c$ is derived from Flesch–Kincaid Grade Level index (FKGL). FKGL~\cite{kincaid1975derivation} refers to the level that must be reached to understand a specific text. Typically, FKGL score is positively correlated to sentence complexity. We normalize the score with the mean and variance calculated from the training data. For complex sentences, $r_c$ is equal to the normalized FKGL, while for simple sentences, $r_c$ = 1 $-$ FKGL, because the model is encouraged to generate sentences with low complexity.

Regard $P_{s \to c}$ and $P_{c \to s}$ as translation policies. Let $\widetilde{x}$ and $\widetilde{y}$ denote the simple and complex sentences obtained by sampling from the current policies. The total reward for sampled sentences can be calculated as:
\begin{align}
    R_c(\widetilde{y}) & = H\left[r_f(\widetilde{y}), r_s(\hat{S}(y), \widetilde{y}) ,r_c(\widetilde{y})\right] \\
    R_s(\widetilde{x}) & = H\left[r_f(\widetilde{x}), r_s(\hat{C}(x), \widetilde{x}) ,r_c(\widetilde{x})\right]
\end{align}
Where $H(\cdot)$ is the harmonic average functions. Comparing with the arithmetic average, the harmonic average can optimize these metrics more equitably. To reduce the variance, sentences obtained by greedy decoding $\hat{x}$ and $\hat{y}$ are used as baselines in the training process:
\begin{align}
    R_s & = R_s(\widetilde{x}) - R_s(\hat{x}) \\
    R_c & = R_c(\widetilde{y}) - R_c(\hat{y})
\end{align}
The loss function is the sum of the negative expected reward for sampled sentence $\widetilde{x}$ and $\widetilde{y}$:
\begin{equation}
    \begin{split}
        \mathcal{L}_{pg}  = - & \mathbb{E}_{\widetilde{x} \sim P_{c \to s}(\cdot|\hat{C}(x))}\left[R_s\right] - \\
        & \mathbb{E}_{\widetilde{y} \sim P_{s \to c}(\cdot|\hat{S}(y))}\left[R_c\right]
    \end{split}
\end{equation}
To optimize this objective function, we estimate the gradient with REINFORCE~\cite{williams1992simple} algorithm:
\begin{equation}
    \begin{split}
        \nabla_{\Theta}\mathcal{L}_{pg} \approx  - & R_s\sum_{i=1}^{|\widetilde{x}|}\log P_{c \to s}(\widetilde{x}_i | \widetilde{x}_{0:i-1}, \hat{C}(x)) - \\
        &  R_c\sum_{i=1}^{|\widetilde{y}|}\log P_{s \to c}(\widetilde{y}_i | \widetilde{y}_{0:i-1}, \hat{S}(y))
    \end{split}
\end{equation}
The final loss is a weighted sum of the cross entropy loss and the policy gradient loss:
\begin{equation}
    \mathcal{L}_{f} = (1 - \gamma)\mathcal{L}_{ce} + \gamma\mathcal{L}_{pg}
\end{equation}
Where $\gamma$ is the parameter to balance the two loss. The complete training process is described in Algorithm \ref{alg:Framwork}.

\section{Experiments}
\subsection{Data}
We use the \textbf{UNTS} dataset~\cite{unsuper-nerual-simplificaiton} to train our unsupervised-model.  The UNTS dataset is extracted from the English Wikipedia dump. It uses automatic metrics\footnote{Mainly by Flesch
Readability Ease} to measure the text complexity and categorize the sentences into complex and simple part. It contains 2M unparalleled sentences.

 \begin{algorithm}[htb]
  \caption{Our Simplification System}
  \label{alg:Framwork}
    \KwIn{Unpaired dataset $\mathcal{S}$, $\mathcal{C}$, Iterations N, parallel data $\mathcal{D}$ (optional).}
    \KwOut{Simplification model $P_{c \to s}$.}
    Train denoising autoencoders $P_{s \to s}$ and $P_{c \to c}$ with $\mathcal{S}$ and $\mathcal{C}$ respectively;

    Initialize the translation model $P_{s \to c}^{(0)}$ and $P_{c \to s}^{(0)}$;

    \For{i=1 to N}{
        Training with parallel data $\mathcal{D}$ (optional);

        Back-translation: Generate simple and complex sentences with current model $P_{c \to s}^{(i-1)}$, $P_{s \to c}^{(i-1)}$;

        Use Generated sentences as training pairs, calculate $\mathcal{L}_{ce}$;

        Sample corresponding sentences with current policy, calculate $\mathcal{L}_{pg}$;

        Update model with $\mathcal{L}_{ce}$ and $\mathcal{L}_{pg}$, get new model $P_{c \to s}^{(i)}$ and $P_{s \to c}^{(i)}$;
        }
    \Return $P_{c \to s}^{(N)}$;
\end{algorithm}
 For semi-supervised training and evaluation, we also use two parallel datasets: \textbf{WikiLarge}~\cite{zhang-lapata-2017-sentence} and \textbf{Newsela} dataset~\cite{xu2015problems}. WikiLarge comprise 359 test sentences, 2000 development sentences and 300k training sentences. Each source sentences in test set has 8 simplified references. Newsela is a corpus extracted from news articles and simplified by professional editors, which is considered to have higher quality and harder than Wiki-Large. Following the settings of ~\citeauthor{zhang-lapata-2017-sentence}~\shortcite{zhang-lapata-2017-sentence}, we discarded the sentence pairs with adjacent complexity. The first 1,070 articles are used for training, next 30 articles for development and others for testing.
\subsection{Training Details}
Our model is built upon Transformer~\cite{vaswani2017attention}. Both encoder and decoders have 3 layers with 8 multi-attention heads. To reduce the vocabulary size and restrict the frequency of unknown words,  we split the words into sub-units with byte-pair encoding (BPE)~\cite{DBLP:conf/acl/SennrichHB16a}. The sub-word embeddings are 512-dimensional vectors pre-trained on the entire data with FastText~\cite{bojanowski2017enriching}. In the training process, we use Adam optimizer~\cite{DBLP:journals/corr/KingmaB14}; the first momentum was set to 0.5 and  batch size to 16. For reinforcement training, we dynamically adjust the balance parameter $\gamma$. At the start of the training process, $\gamma$ is set to zero, which can help model converge rapidly and shrink the search space. As training progresses, $\gamma$ is gradually increased and finally converge to 0.9. We use the sigmoid function to perform this process.

The system is trained in both unsupervised and semi-supervised manner. We pre-train the asymmetric denoising autoencoders for 200,000 steps with a learning rate of 1e-4, After that, we add back-translation training with a learning rate of 5e-5. As for semi-supervised training, we randomly select 10\% data from the corresponding parallel corpus and the model is trained alternately between denoising autoencoders, back-translation, and parallel sentences.

\subsection{Metrics and Model Selection}
Following the previous studies, we use corpus level SARI~\cite{xu2016optimizing} as our major metrics. SARI measures whether a system output can correctly keep, delete and add from the complex sentence. It calculates the N-gram overlap of these three aspects between system outputs and reference sentences. SARI is the arithmetic mean of F1-scores of three rewrite operations\footnote{For corpus level SARI, the original script provided by \citeauthor{xu2016optimizing}~\shortcite{xu2016optimizing} is only for 8 references WikiLarge dataset. We confirmed this fact with the author. So in our experiments, we use the original script for WikiLarge corpus and our own script for 1 reference Newsela corpus. Several previous works misused the original scripts on the 1 reference dataset which may lead to a very low score.}.  We also use BLEU score as an auxiliary metric. Although BLEU is reported to have a negative correlation with simplicity~\cite{DBLP:conf/emnlp/SulemAR18}, it often positively correlates with grammaticality and adequacy. This may help us give a comprehensive evaluation for different systems.

For model selection, we mainly use SARI to tune our model. However, SARI rewards deletion, which means large differences may lead to good SARI even though the output is ungrammatical or irrelevant. To tackle this problem, we introduce BLEU score threshold similar to \citeauthor{vu-etal-2018-sentence}~\shortcite{vu-etal-2018-sentence}. epochs with BLEU score lower than threshold $\xi$ will be ignored. We set $\xi$ to 20 on Newsela dataset and 70 on Wiki-Large dataset.

\subsection{Comparing Systems and Model Variants}
We compare our system with several baselines. For unsupervised model, we considered UNTS\cite{unsuper-nerual-simplificaiton} ---- a neural encoder-decoder model based on adversarial training; and a rule-based lexical simplification system called LIGHT-LS~\cite{glavavs2015simplifying}.
 Multiple supervised systems are also used as baselines, including Hybrid~\cite{narayan2014hybrid}, PBMT-R~\cite{wubben2012sentence} and DRESS\footnote{The system outputs of PBMT-R, Hybrid, and
DRESS are publicly available.}~\cite{zhang-lapata-2017-sentence}. We also trained a Seq2Seq model based on vanilla Transformer.

Using our approach, we also propose three different variants for experiments. (1) Basic Back-Translation based unsupervised TS model (BTTS). (2) Model integrated with reinforcement learning (BTTSRL). (3) Semi-Supervised model with limited supervision using 10\% labelled data (BTTS+10\%) and with full supervision using all labelled data (BTTS+full).

\section{Results}
In this section, we present the comparison results on both standard automatic evaluation and human evaluation. We also analyze the effects of different noising type in back-translation with ablation study.

\begin{table}[htbp]
	\centering
		\resizebox{\columnwidth}{!}{
		\begin{tabular}{l|c|ccc|c}
		\toprule
          \textbf{Newsela\quad} & SARI & \multicolumn{3}{|l|}{Component of SARI} & BLEU \\
          & & $F_{keep}$ & $F_{del}$ & $F_{add}$ &   \\
          \midrule
          \multicolumn{6}{l}{\textbf{Supervised Model}} \\
          PBMT-R& 26.95 & 37.63 & 39.38 & 3.84 &  18.10 \\
          Hybrid& 35.09 & 30.08 & 73.48 & 1.71 & 14.45  \\
          Seq2Seq(10\%)& 34.16 & 14.86 & \textbf{86.26} & 1.34 & 2.34   \\
          Seq2Seq(full)& 39.18 & 34.26 & 80.76 & 2.52 & 12.50   \\
          DRESS & 38.91 & 41.37 & 71.52 & 3.84 & 23.19 \\
          DRESS-LS & 38.63 & \textbf{41.97} & 69.91 & 4.02 & \textbf{24.25}  \\
        \midrule
        \multicolumn{6}{l}{\textbf{Unsupervised Model}} \\
          LIGHT-LS& 28.41 & 35.70 & 48.47 & 1.08 & 14.69  \\
          UNTS& 34.01 & 39.98 & 60.15 & 1.90 & 19.26  \\
          BTTS& 37.07 & 38.76 & 69.55 & 2.91 & 19.67 \\
          BTRLTS & 37.69 & 37.15 & 73.22 & 2.71 & 19.55  \\
        \midrule
        \multicolumn{6}{l}{\textbf{Semi-supervised Model}} \\
          BTTS(+10\%)& 38.69 & 40.88 & 71.13 & 4.07 & 20.89  \\
          BTRLTS(+10\%)& 38.62 & 40.34 & 71.81  & 3.72 & 21.48  \\
          BTTS(+full)& \textbf{39.69}& 40.75 & 73.89 & \textbf{4.26} & 23.05  \\
        \bottomrule
        \multicolumn{6}{l}{}
        \end{tabular}
		}

	\qquad
		\centering
		\resizebox{\columnwidth}{!}{
		    \begin{tabular}{l|c|ccc|c}
       \toprule
          \textbf{WikiLarge\quad} & SARI & \multicolumn{3}{|l|}{Component of SARI} & BLEU \\
          & & $F_{keep}$ & $F_{del}$ & $F_{add}$ &    \\
          \midrule
          \multicolumn{6}{l}{\textbf{Supervised Model}} \\
          PBMT-R& \textbf{38.55} & 73.03 & 36.90 & \textbf{5.72} &  81.05 \\
          Hybrid& 31.40 & 46.87 & \textbf{45.48} & 1.84 & 48.67  \\
          Seq2Seq(10\%)& 28.06 & 30.71 & 50.96 & 2.51 & 19.86   \\
          Seq2Seq(full)& 36.53 & \textbf{74.48} & 32.35 & 2.79 &  \textbf{87.75} \\
          DRESS & 37.08 & 65.16 & 43.13 & 2.94 & 77.35  \\
          DRESS-LS & 37.27 & 66.78 & 42.19 & 2.81 & 80.38  \\
        \midrule
        \multicolumn{6}{l}{\textbf{Unsupervised Model}} \\
          LIGHT-LS& 35.12 & 63.93 & 39.61 & 1.81 & 59.69  \\
          UNTS& 37.04 & 65.21 & 44.33 & 1.60 & 74.54  \\
          BTTS& 36.98 & 70.88 & 37.46 & 2.59 & 78.36  \\
          BTRLTS & 37.06 & 70.44 & 37.96 & 2.78 & 77.37   \\
        \midrule
        \multicolumn{6}{l}{\textbf{Semi-supervised Model}} \\
          BTTS(+10\%)& 37.25 & 68.82 & 40.06 & 2.87 & 80.06  \\
          BTRLTS(+10\%)& 37.00& 71.41 & 36.62 & 2.97 & 83.39  \\
          BTTS(+full)& 36.92 & 72.79 & 35.93 & 2.04 & 87.09  \\
        \bottomrule
        \end{tabular}
		}
	\caption{Results on Newsela and Wiki-Large dataset}\label{tab:main_report}
\end{table}

\subsection{Automatic Evaluation}
We report the results in Table \ref{tab:main_report}. For unsupervised systems, our model outperforms previous unsupervised baselines on both datasets. Compared to LIGHT-LS and UNTS, our model achieves a large improvement (+9.28, +3.68 on SARI) on Newsela dataset. On Wiki-Large dataset, our model still outperforms the LIGHT-LS and gets similar results with UNTS on SARI, but achieves higher BLEU score. This means our model can generate more fluently and the outputs are
 more relevant to the source sentences. Furthermore, our unsupervised models perform closely to the state-of-the-art supervised systems. The results also show that reinforcement training is helpful to unsupervised systems. It brings 0.62 SARI  improvement on Newsela and 0.08 on Wiki-Large corpus.

In addition, the results of the semi-supervised systems show that our model can greatly benefit from small amounts of parallel data. Model trained with 10\% of the parallel sentences can perform competitively with state-of-the-art supervised systems on both datasets. With the increase of parallel data, all metrics can be further improved on Newsela corpus. Semi-supervised model trained with full parallel sentences significantly outperform the state-of-the-art TS models such as DRESS-LS (+1.03 SARI). On Wiki-Large dataset, the BLEU score has 3.7 point improvement with the full parallel sentences, but we cannot observe any improvements on SARI metrics. This might because the simplified sentences in Wiki-Large are often too closed to the source sentences or even not simpler at all~\cite{xu2015problems}. This defect may motivate the system to copy directly from the source sentences, which cause a decline on SARI score.

Both unsupervised and semi-supervised model achieve better improvement on Newsela dataset, showing that by leveraging large amount of unpaired data, our models can learn simplification better on harder and smaller datasets.

\begin{table}[htbp]
		\centering
		\resizebox{\columnwidth}{!}{
		\begin{tabular}{l|llll}
		\toprule
          \textbf{Newsela\quad} & Fluency & Adequacy & Simplicity & Avg. \\
          \midrule
          Hybrid & 2.35** & \ \ 2.37** & \ \ 2.61 & 2.44* \\
          Seq2Seq & 2.26** & \ \ 1.82** & \ \ 2.34* & 2.14** \\
          DRESS & \textbf{3.13} & \ \ 2.72 & \ \ 2.71 & 2.85 \\
          BTTS(+10\%) & 3.12 & \ \ \textbf{2.87} & \ \ 2.58 & \textbf{2.86} \\
          BTTS(+full) & 3.08 & \ \ 2.59 & \ \ \textbf{2.85}* & 2.84 \\
        \midrule
          Reference & 3.62** & \ \ 2.85 & \ \ 3.17** & 3.21* \\
        \bottomrule
        \multicolumn{5}{l}{}
        \end{tabular}
		}
	\qquad
    	\resizebox{\columnwidth}{!}{
    	\begin{tabular}{l|llll}
    	\toprule
          \textbf{WikiLarge\quad} & Fluency & Adequacy & Simplicity & Avg. \\
          \midrule
          Hybrid & \ \ 2.98* & \ \ 2.68** & \ \ \textbf{3.0}* & 2.89 \\
          Seq2Seq & \ \ 3.13 & \ \ 3.24 & \ \ 2.75 & 3.04 \\
          DRESS & \ \ \textbf{3.43} &\ \  3.15 & \ \ 2.86 & \textbf{3.15} \\
          BTTS(+10\%) & \ \ 3.39 & \ \ 3.35 & \ \ 2.59 & 3.11 \\
          BTTS(+full) & \ \ 3.42 & \ \ \textbf{3.36} & \ \ 2.61 & 3.13 \\
        \midrule
          Reference & 3.42 & 3.34 & 3.02* & 3.26 \\
        \bottomrule
        \end{tabular}
    	}
	\caption{Human evaluation on Newsela and Wiki-Large. Ratings significantly different from our model are marked with ${*}$ ($p < 0.05$) and ${**}$ ($p < 0.01$). We use student t-test to perform significance tests}\label{tab:human_report}
\end{table}

\begin{table*}[h]
    \centering
    \begin{tabular}[width=0.9\columnwidth]{l|cccc|cccc}
    \toprule
           & \multicolumn{4}{c}{\textbf{Newsela}} & \multicolumn{4}{|c}{\textbf{WikiLarge}} \\
    \midrule
          \textbf{Noise Type} & \textbf{SARI} & $F_{keep}$ & $F_{del}$ & $F_{add}$ & \textbf{SARI} & $F_{keep}$ & $F_{del}$ & $F_{add}$ \\
    \midrule
          original(drop \& shuffle) & 33.63 & 38.60 & 59.28 & 3.01 & 35.91& 70.74 & 34.97& 2.03 \\
          \quad + additive &35.94 & 39.10 & 65.11 & 3.59 & 36.85& \textbf{70.93} & 36.99 & 2.63\\
          \quad\quad + substitution & \textbf{38.69} & \textbf{40.88} & \textbf{71.13} & \textbf{4.07} & \textbf{37.25} & 68.82 & \textbf{40.06} & \textbf{2.87}\\
    \bottomrule
    \end{tabular}
    \caption{SARI score of models with different noise. All models are trained in semi-supervised manner with 10\% parallel corpus.}
    \label{tab:abli_report}
\end{table*}

\subsection{Human Evaluation}
Due to the limitations in automatic metrics, we also conduct human evaluation on two datasets. We randomly select 200 sentences generated by our systems and the baselines as test samples. Similar to previous work~\cite{zhang-lapata-2017-sentence}, we ask native English speakers to evaluate the fluency, adequacy, and simplicity of the test samples via Amazon Mechanical Turk. The three aspects are rated on a 5-point Likert Scale. We use our semi-supervised model to perform human evaluation. The results are illustrated in Table \ref{tab:human_report}.

On Newsela dataset, our model gets comparable results with DRESS and substantially outperforms Hybrid and fully supervised sequence-to-sequence model. Although sequence-to-sequence model has obtained promising scores on SARI (see in Tab \ref{tab:main_report}), it performs the worst on adequacy and rather poor on fluency. This also proved that SARI only weakly correlates with judgments on fluency and adequacy~\cite{xu2016optimizing}. We have similar results on Wiki-Large dataset and our model achieves the highest score on adequacy.
\subsection{Ablation Study}
We perform ablation study to analyze the effects of denoising type on simplification performance. We test three types of noise:
\begin{itemize}
    \item[a.] Original noise in machine translation including word dropout and shuffling (denoted as original)
    \item[b.] Original noise plus with additive noise on simple sentences.
    \item[c.] Substitution noise introduced on top of (b), which is our proposed noise type above.
\end{itemize}

\begin{figure}[htb]
    \centering
    \includegraphics[width=0.95\columnwidth]{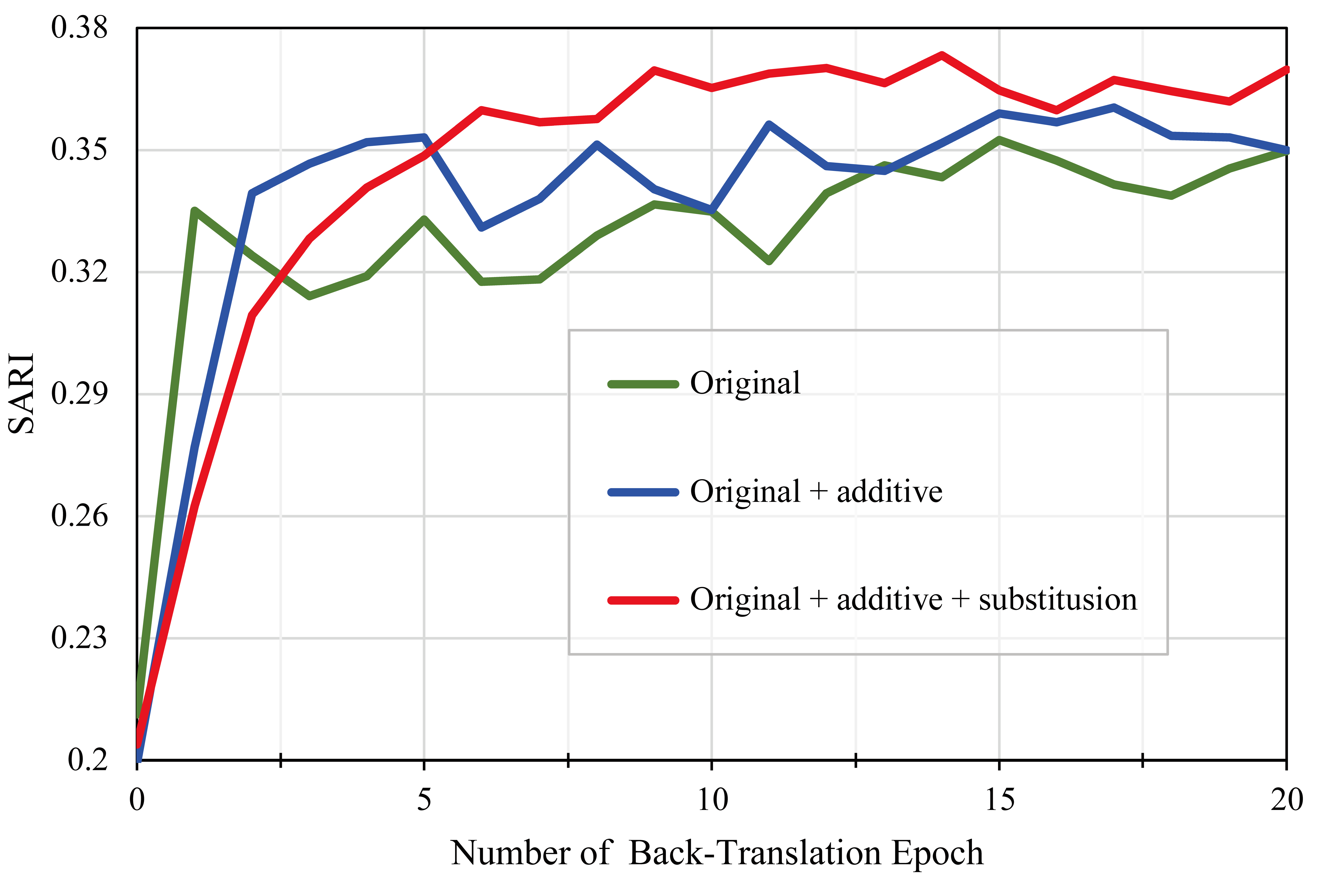}
    \caption{SARI variation in semi-supervised training process
    under different types of noise.
    }
    \label{fig:abli_noise}
\end{figure}

Note that denoising autoencoders with different noise type may have varied convergence rate. To make a better comparison, we pre-train these autoencoders with different steps until they reach similar training loss. In our experiment, we pre-train 20,000 steps for autoencoders with noise type (a), 50,000 steps for noise type (b) and 200,000 steps for noise type (c). Figure \ref{fig:abli_noise} shows the variation of SARI on the development set with the change of back-translation epoch in semi-supervised training. The model with only word dropout and shuffle remains at low scores during the training process, while our proposed model has made a significant improvement.

Furthermore, we analyze the insights of SARI score in detail. Table \ref{tab:abli_report} illustrate SARI score and its components under different types of noise. Additive noise in simple sentences can significantly promote the delete and add operation. Substitution also has a similar effect and makes a further improvement. Model with original noise tend to copy directly from the source sentence, resulting a relative higher F-score in keep operation, but much lower scores on other aspects.

\section{Conclusions}
In this paper, we adopt back-translation architecture to perform unsupervised and semi-supervised text simplification. We propose a novel asymmetric denoising autoencoder to model simple and complex corpus separately, which can help the system learn structures and features from the sentence with different complexity. Ablation study demonstrates that our proposed noise type can significantly promote the system performance comparing with basic denoising method. We also integrate reinforcement learning and achieve better SARI score on unsupervised models. Automatic evaluation and human judgment show that with limited supervision, our model can perform competitively with multiple full supervised systems. We also find the automatic metrics cannot correlate well with the human evaluation. We
plan to investigate a better method in future work.
\section{Acknowledgments}
This work has been supported by the National Key Research and Development Program of China (Grant No.2017YFB1002102) and the  China NSFC projects (No.61573241). We thank the anonymous reviewers for their thoughtful comments and efforts towards improving  this manuscript.

\bibliography{AAAI-ZhaoY.8044}
\bibliographystyle{aaai}
\end{document}